\documentclass[11pt]{article}

\usepackage[T1]{fontenc}
\usepackage[utf8]{inputenc}
\usepackage{newtxtext,newtxmath}
\usepackage{fvextra}
\usepackage{geometry}
\usepackage{microtype}
\usepackage{graphicx}
\usepackage{booktabs}
\usepackage{caption}
\usepackage{titlesec}

\usepackage[dvipsnames,table]{xcolor}

\usepackage[most]{tcolorbox}

\tcbset{
  titlecard/.style={
    enhanced, frame hidden, boxrule=0pt,
    colback=cardbg, arc=12pt,
    left=14pt, right=14pt, top=14pt, bottom=14pt
  }
}

\usepackage{tikz}
\usepackage{adjustbox}   
\usepackage{array}
\usepackage{makecell}
\usepackage{multirow}    
\usepackage{enumitem}

\usepackage[authoryear,round,sort]{natbib}
\bibpunct{(}{)}{;}{a}{,}{,}

\usepackage[colorlinks=true]{hyperref}
\hypersetup{
  linkcolor=blue!60!black,  
  citecolor=blue!60!black,  
  urlcolor=blue!60!black    
}

\usepackage[table]{xcolor}
\usepackage[most]{tcolorbox}  
\usepackage{xstring}
\usepackage{amsmath}          

\definecolor{myBlue}{RGB}{62,111,210}

\tcbset{
  pill/.style={
    on line, boxsep=0.25mm, left=0.6mm, right=0.6mm, top=0.1mm, bottom=0.1mm,
    colframe=black!0, arc=1.5pt
  },
  card/.style={
    enhanced, breakable,
    colback=white, colframe=myBlue, boxrule=1.2pt,
    arc=8pt, outer arc=8pt,
    left=3mm, right=3mm, top=2mm, bottom=2mm,
    colbacktitle=myBlue, coltitle=white,
    fonttitle=\bfseries\large,
    toptitle=1mm, bottomtitle=1mm
  }
}

\newtcbox{\chgpos}[1][]{pill, colback=blue!12, #1}
\newtcbox{\chgneg}[1][]{pill, colback=red!12,  #1}
\newcommand{\chg}[1]{%
  \ifmmode \text{%
    \IfBeginWith{#1}{-}{\chgneg{\scriptsize$\downarrow$\,\StrGobbleLeft{#1}{1}}}{%
    \IfBeginWith{#1}{+}{\chgpos{\scriptsize$\uparrow$\,\StrGobbleLeft{#1}{1}}}{%
                         \chgpos{\scriptsize$\uparrow$\,#1}}}}%
  \else
    \IfBeginWith{#1}{-}{\chgneg{\scriptsize$\downarrow$\,\StrGobbleLeft{#1}{1}}}{%
    \IfBeginWith{#1}{+}{\chgpos{\scriptsize$\uparrow$\,\StrGobbleLeft{#1}{1}}}{%
                         \chgpos{\scriptsize$\uparrow$\,#1}}}%
  \fi
}

\newtcolorbox{policybox}[2][]{card, title={#2}, #1}

\usepackage{xcolor}
\definecolor{cardbg}{RGB}{245,246,248} 
\usepackage[most]{tcolorbox}
\tcbset{
  titlecard/.style={
    enhanced, frame hidden, boxrule=0pt,
    colback=cardbg, arc=12pt,
    left=14pt, right=14pt, top=14pt, bottom=14pt
  }
}

\usepackage[dvipsnames]{xcolor}
\AtBeginDocument{\hypersetup{
  colorlinks=true,
  linkcolor=blue!60!black,
  citecolor=blue!60!black,  
  urlcolor=blue!60!black
}}

\usepackage[authoryear,round,sort]{natbib}
\bibpunct{(}{)}{;}{a}{,}{,}   

\definecolor{orange1}{RGB}{255, 204, 153}

\definecolor{green1}{RGB}{209, 226, 255}
\definecolor{green2}{RGB}{163, 197, 255}
\definecolor{green3}{RGB}{121, 170, 255}

\usepackage{pifont}
\newcommand{\cmark}{\textcolor[RGB]{0,100,0}{\ding{51}}}

\newcommand{\xmark}{\textcolor{red}{\ding{55}}}

\usepackage{tikz}
\usepackage{adjustbox} 

\newcommand{\LogoPathA}{}   
\newcommand{\LogoPathB}{}  
\newcommand{\LogoHeight}{11mm}          
\newcommand{\LogoMaxWidth}{26mm}        
\newcommand{\LogoGap}{4mm}              

\newcommand{\PaperTitle}{Remedy-R: Generative Reasoning for Machine Translation Evaluation without Error Annotations\vspace{1mm}}
\newcommand{\AuthorsTop}{%
  \hspace{-2mm}\textbf{\small Shaomu Tan}$^{\diamondsuit,\heartsuit,*}$\hspace{0.2em}
  \textbf{\small Ryosuke Mitani}$^{\heartsuit}$\hspace{0.2em}
  \textbf{\small Ritvik Choudhary}$^{\heartsuit}$\hspace{0.2em}
  \textbf{\small Qiyu Wu}$^{\heartsuit}$\hspace{0.2em}
  \textbf{\small Toshiyuki Sekiya}$^{\heartsuit}$\hspace{0.2em}
  \textbf{\small Christof Monz}$^{\diamondsuit}$%
}
\newcommand{\Affiliations}{%
  \vspace{-3mm} $^{\diamondsuit}$ \small University of Amsterdam\qquad
  $^{\heartsuit}$ \small Sony Group Corporation \qquad $^{*}$\ Work done while interning at Sony.%
}
\newcommand{\Contact}{}

\newcommand{\AbstractText}{%

Over the years, automatic MT metrics hillclimbed benchmarks and presented strong and sometimes human-level agreement with human ratings. Yet they remain black-box, offering little insight into their decision-making and often failing under real-world out-of-distribution (OOD) inputs. We introduce Remedy-R, a reasoning-driven generative MT metric trained with reinforcement learning from pairwise translation preferences, without requiring error-span annotations or distillation from closed LLMs. Remedy-R produces step-by-step analyses of accuracy, fluency, and completeness, followed by a final score, enabling more interpretable assessments. With only 60K training pairs across two language pairs, Remedy-R remains competitive with top scalar metrics and GPT-4-based judges on WMT22–24 meta-evaluation, generalizes to other languages, and exhibits strong robustness on OOD stress tests. Moreover, Remedy-R models generate self-reflective feedback that can be reused for translation improvement. Building on this finding, we introduce Remedy-R Agent, a simple evaluate–revise pipeline that leverages Remedy-R’s evaluation analysis to refine translations. This agent consistently improves translation quality across diverse models, including Qwen2.5, ALMA-R, GPT-4o-mini, and Gemini-2.0-Flash, suggesting that Remedy-R’s reasoning captures translation-relevant information and is practically useful.

}

\makeatletter
\renewcommand{\maketitle}{%
  \begingroup\setlength{\parindent}{0pt}%
  \begin{center}
    \begin{tcolorbox}[
      titlecard, before upper=\vspace{0pt}, 
      overlay={
  \node[anchor=south east, inner sep=6pt] at (frame.south east){%
    \begingroup
    \edef\tempA{\LogoPathA}\edef\tempB{\LogoPathB}%
    \ifx\tempA\empty\else
      \IfFileExists{\LogoPathA}{%
        \adjincludegraphics[max height=\LogoHeight, max width=\LogoMaxWidth, keepaspectratio]{\LogoPathA}%
      }{}%
    \fi
    \ifx\tempB\empty\else
      \hspace{\LogoGap}%
      \IfFileExists{\LogoPathB}{%
        \adjincludegraphics[max height=\LogoHeight, max width=\LogoMaxWidth, keepaspectratio]{\LogoPathB}%
      }{}%
    \fi
    \endgroup
  };
}
    ]
      {\Huge\bfseries \PaperTitle \par}
      \vspace{0.9em}
      {\large
        \begin{tabular}{c}
          \AuthorsTop \\[0.45em]
        \end{tabular}\par}
      \vspace{0.6em}
      {\normalsize \Affiliations \par}
      \vspace{0.25em}
      {\ttfamily \Contact \par}

      \vspace{0.9em}
      {\normalsize\bfseries \par}
      \vspace{0.35em}
      \begin{minipage}{1.0\linewidth}\small
        \AbstractText
      \end{minipage}

      \vspace{0.8em}
      \noindent\makebox[\linewidth][l]{\rule{0.3\linewidth}{0.7pt}}%
      \par\vspace{0.4em} 
      \begin{minipage}{0.96\linewidth}
        \footnotesize
        \textbf{Date:} October 15, 2025\\
        \textbf{Correspondence:} \href{mailto:s.tan@uva.nl}{s.tan@uva.nl}\\
        \textbf{Project:} \href{https://github.com/Smu-Tan/Remedy-R}{github.com/Smu-Tan/Remedy-R}\\
        \textbf{Models:} \href{https://huggingface.co/ShaomuTan/Remedy-R-14B}{huggingface.co/ShaomuTan/Remedy-R-14B}
      \end{minipage}
    \end{tcolorbox}
  \end{center}
  \endgroup
}
\makeatother

\usepackage{CJKutf8}
\AtBeginDocument{\begin{CJK*}{UTF8}{gbsn}} 
\AtEndDocument{\end{CJK*}}
\begin{document}
\maketitle

\section{Introduction}

Automatic evaluation metrics for machine translation (MT) have evolved from string-matching metrics such as BLEU~\citep{papineni2002bleu} and ChrF~\cite{popovic2015chrf} to learned neural metrics. Systems like xCOMET~\citep{guerreiro2024xcomet}, MetricX~\citep{juraska2024metricx}, and Remedy~\citep{tan2025remedy} achieve strong correlations with human preferences, and in some settings report agreement that even exceeds expert annotators~\citep{proietti2025has}. Despite these advances, existing metrics act as black boxes: they assign a single scalar score without explaining why, obscuring the decision-making process and offering no insight into what makes a translation good or bad. This lack of explainability means that even when a metric aligns with human ratings, its underlying decision-making remains invisible.\\

\noindent Moreover, translation quality is multi-dimensional, involving factors such as accuracy, fluency, and completeness. A single number cannot reveal which dimensions influenced the judgment, making metric outputs difficult to interpret and diagnose, for example when meaning is accurate but fluency is poor. Crucially, explainability is not only desirable for transparency but also essential for robustness. Without explicit reasoning, metrics may rely on spurious cues learned during training, which we later show leads to failures under out of distribution (OOD) conditions such as source copy and input perturbations~\citep{lo2023metric,knowles2024mslc24,moghe2025machine}. These shortcomings make today’s metrics powerful but opaque, limiting their reliability and their usefulness as diagnostic or interactive evaluation tools.\\



\noindent Recent efforts have attempted to enhance explainability through error-span prediction, where models highlight the words or phrases that cause translation errors~\citep{guerreiro2024xcomet,treviso2024xtower}. Such methods provide a form of local interpretability by showing which parts of the translation are likely incorrect. However, span-based feedback is inherently limited: it struggles with omissions and discourse-level coherence, so an explanation can look precise yet still fail to reflect overall adequacy. Consequently, the explanation may appear precise but still fail to reflect overall translation adequacy.\\


\noindent A promising direction is to use LLMs-as-judges to perform quality analysis and evaluate translation across multiple dimensions. However, many existing approaches~\citep{kocmi2023large,fernandes2023devil,qian2024large} simply ask the model to directly predict a quality score, essentially remaining black boxes. More recent systems~\citep{lu2024error,kocmi2023gemba,zhang2025himate,feng2025mad} attempt to simulate human annotators by performing multi-dimensional error analysis under the MQM framework~\citep{freitag2021experts}. Yet, they rely on prompting closed commercial LLMs, and open models struggle to match their evaluation performance~\citep{lu2025mqm}. This raises an open question: how can we train open reasoning-based metrics that can analyze translation quality and align their judgments with human preferences?\\

\begin{policybox}[width=\columnwidth]{Remedy-R ref-free Evaluation Example}
\footnotesize
\textbf{Source}: Even if I'm not in the income level that would pay more taxes, I can still buy fewer things that are more expensive so that they can be made by people making a living wage and I can be willing to wait a few days to get it so that some gig worker doesn't have to be worked to the bone...\\[5pt]
\textbf{Translation}: Sogar wenn ich nicht in der Einkommensklasse bin, in der mehr Steuern gezahlt werden, kann ich immer noch weniger teure Dinge kaufen, sodass sie von Menschen hergestellt werden können, die maßgeblich dazu beitragen, und ich bin bereit, ein paar Tage zu warten, bevor der Arbeitslohn abläuft…\\[5pt]
\textbf{Remedy-R's Evaluation}: \\Let's evaluate the translation step by step:

\begin{itemize}[leftmargin=*,itemsep=4pt]
  \item \textbf{Accuracy}
  \begin{itemize}[leftmargin=1.5em,nosep]
    \item "sodass sie von Menschen hergestellt werden können, die maßgeblich dazu beitragen" is not clear and does not accurately reflect the original meaning. The original text states that the products should be made by people who are paid a living wage, but the translation does not convey this clearly.
    \item "bevor der Arbeitslohn abläuft" is incorrect and does not make sense in this context. The original text states "so that some gig worker doesn't have to be worked to the bone" which means the gig worker is not overworked. The translation should reflect this meaning.
  \end{itemize}

  \item \textbf{Fluency}
  \begin{itemize}[leftmargin=1.5em,nosep]
    \item The sentence structure is somewhat complex and could be clearer.
    \item There are some awkward phrasings, such as "in der Einkommensklasse bin, in der mehr Steuern gezahlt werden," which could be simplified.
    \item The use of "maßgeblich dazu beitragen" is not clear and does not fit the context.
  \end{itemize}

  \item \textbf{Completeness}
  \begin{itemize}[leftmargin=1.5em,nosep]
    \item The original text also states that the speaker is willing to wait a few days to get the product so that a gig worker doesn't have to be overworked. This part is also not clearly conveyed in the translation.
  \end{itemize}
\end{itemize}

\vspace{4pt}
\noindent\#\#\#\# Score: 65.
\end{policybox}\label{box:example}

\vspace{5mm}\noindent In this paper, we introduce \textit{Remedy-R}, a reasoning MT metric that produces step-by-step analyses along three dimensions (accuracy, fluency, completeness) and a final quality score (see Example~\ref{box:example}). Inspired by the pairwise preference framework in Remedy~\citep{tan2025remedy}, Remedy-R is trained with Reinforcement Learning via Verifiable Rewards (RLVR) using \textit{pairwise human preferences}. Specifically, we define a deterministic verifiable reward function that checks whether the model-implied preference (from parsed final scores) matches the human preference label. Remedy-R is then optimized with PPO~\citep{schulman2017proximal} to maximize these verifiable rewards. Unlike xTower that depends on GPT-4 distillation and supervised fine-tuning with error spans~\citep{treviso2024xtower}, Remedy-R starts training directly with RLVR on MT preference data, without requiring error span annotations or SFT cold-start~\citep{guo2025deepseek}.\\

\noindent To assess whether Remedy-R’s reasoning explanations are faithful and practically useful, we take two complementary approaches: (i) we validated the faithfulness of Remedy-R’s explanations with GPT-4o-mini (§\ref{sec:agent-faithfulness}), and (ii) we build \textit{Remedy-R Agent}, a simple evaluate-revise loop that uses Remedy-R’s feedback to refine translations (§\ref{sec:agent-1}); we use this loop as a probe for rationale utility rather than as a new translation agent framework. Despite never being trained for post-editing, the agent consistently improves translation quality across both high- and low-resource languages and across outputs from diverse MT systems and LLMs (Qwen2.5~\citep{Yang2024Qwen25TR}, ALMA-R~\citep{xu2024contrastive}, GPT-4o-mini~\citep{achiam2023gpt}, and Gemini-2.0-Flash) on WMT24++~\citep{kocmi2024findings,deutsch2025wmt24++}. Unlike xTower, whose generalization is limited to its training languages, Remedy-R remains effective across 11 language pairs while trained on only two. Our contributions are as follows:

\begin{itemize}
\item We propose \textit{Remedy-R}, a generative MT evaluator that produces step-by-step analyses and a final score, trained directly via RLVR without error span annotation or distillation.\smallskip
\item Remedy-R achieves competitive performance on WMT22–24 meta-evaluation with only 60K training pairs from two language pairs, and exhibits strong OOD behavior.\smallskip
\item We introduce Remedy-R Agent as a lightweight probe for explanation utility: treating rationales as feedback consistently improves translations without post-editing training.\smallskip
\item Remedy-R Agent improves translations across Qwen2.5, ALMA-R, GPT-4o-mini, Gemini-2.0-Flash, and maintains cross-lingual generalization beyond its training languages.
\end{itemize}

\section{Method}

\subsection{Task Formulation}
\label{sec:task-formulation}

We revisit MT evaluation as follows. Given a source sentence $\mathit{src}$, a translation $\mathit{mt}$, and an optional reference $\mathit{ref}^*$ (with $\mathit{ref}^*=\emptyset$ in reference-free settings), a metric $M$ outputs a scalar quality score. Conventional learned metrics directly map $(\mathit{src}, \mathit{mt}, \mathit{ref}^*)$ to a single number, which limits interpretability and reusability for downstream refinement. In Remedy-R, we formulate MT evaluation as \textit{conditional text generation} guided by an instruction that specifies the task and criteria, namely accuracy, fluency, and completeness. The model follows a reason then score protocol: it first writes a short analysis and then outputs a numeric score that we can parse for evaluation. Here, let the input be:
\[
\mathbf{x} \;=\; \langle \mathit{inst},\ \mathit{src},\ \mathit{mt},\ \mathit{ref}^* \rangle.
\]
where the instruction $\mathit{inst}$ specifies the evaluation task and criteria (e.g., accuracy, fluency, completeness). The model produces an output sequence $\mathbf{y}$ in a \emph{reason-then-score} format:  a step-by-step reasoning analysis and a final numeric score in $[0,100]$. We parameterize a conditional policy $\pi_\theta(\mathbf{y}\mid \mathbf{x})$ and generate autoregressively:
\begin{equation}
\pi_\theta(\mathbf{y}\mid \mathbf{x}) \;=\; \prod_{t=1}^{T} \pi_\theta\!\big(y_t \mid \mathbf{x}, y_{<t}\big).
\label{eq:autoregressive}
\end{equation}
Here $T$ denotes the output length. The final score in the output is directly parsed for evaluation and will support the verifiable rewards used in §\ref{sec:reward-design}. 

\subsection{Remedy\texorpdfstring{-}{-}R Reward Design}
\label{sec:reward-design}

We optimize Remedy-R with reinforcement learning using a reward that is simple to verify and closely aligned with human judgments. The reward has two components: (i) a sparse \emph{pairwise ranking} signal that enforces the correct quality ranking of two translations, aligning the model’s preferences with human judgments; and (ii) a \emph{reward shaping} term that 
shapes the model’s predicted numeric scores toward human scores, turning the sparse signal into a richer, continuous one. The RL algorithm (PPO) maximizes the expected reward defined below; see §\ref{sec:rlvr-ppo} for details.

\subsubsection{Pairwise ranking reward.}
For each source sentence, we have two translations $\mathit{mt}_A$ and $\mathit{mt}_B$ with human-annotated scores $g_A,g_B\in[0,100]$. We determine which translation is better by simply comparing $g_A$ and $g_B$ and use this as the ground truth label; pairs with $g_A = g_B$ (ties) are excluded from training. We then instruct the model to produce a reasoning COTs path and two final quality scores $s_A,s_B\in[0,100]$ in a fixed format (see the training template below). Instead of predicting a ranking label directly, we ask the model to evaluate A and B independently and to assign scores to each; this enables single segment evaluation at inference time without quadratic pairwise comparisons. Lastly, we randomize the A or B order during training to reduce position bias~\citep{sproat2025transevalnia}.

\noindent Such structured format allows us to directly extract the predicted scores $s_A$ and $s_B$ from the model’s output. Some prior work enforces strict output formatting with special tokens and format rewards, such as \texttt{<think>}~\citep{guo2025deepseek}. We found that our simpler format can be learned easily, so additional format rewards are unnecessary. \\

\noindent The pairwise ranking reward then checks whether the model’s predicted ranking matches the human ranking:
\begin{equation}
r_{\text{rank}} =
\begin{cases}
1, & \text{if model and human rankings agree},\\
0, & \text{otherwise}.
\end{cases}
\label{eq:reward-rank}
\end{equation}
This ranking reward is sparse and binary, but fully verifiable because it depends only on the parsed score block and human labels. If the score block cannot be parsed or falls outside the valid range, the reward is zero, and ties are treated as zero reward as well.

\subsubsection{Reward shaping.}

The pairwise ranking reward ensures that the model learns to rank translations correctly. However, the sparse binary pairwise ranking reward provides no information about how good each translation is. For instance, if the model outputs $(s_A, s_B) = (100, 99)$ and the human scores are $(g_A, g_B) = (100, 0)$, both cases receive $r_{\text{rank}} = 1$ despite very different scoring behavior. To provide richer feedback, we add a reward shaping term that measures how close the predicted scores are to the human scores, making the overall reward more informative and dense. The shaping term provides score calibration beyond ranking, making the feedback denser while keeping the ranking signal verifiable.\\

\noindent Specifically, we adopt a Huber-style penalty for this shaping term. Considering human evaluation scores are often noisy and inconsistent, especially when multiple annotators are involved or when translations are of similar quality~\citep{freitag2021experts, tan2025remedy}, the Huber penalty offers a good balance: small errors within a threshold $c$ are penalized quadratically and smoothly, while larger errors receive a linearly increasing penalty, making the reward robust to annotation noise but still sensitive to large outliers. Specifically, for per-candidate errors $e_A = s_A - g_A$ and $e_B = s_B - g_B$, the Huber penalty is defined as:

\begin{equation}
\rho_c(e) =
\begin{cases}
\frac{1}{2}\frac{e^2}{c}, & |e| \le c,\\[2pt]
|e| - \frac{1}{2}c, & |e| > c,
\end{cases}
\label{eq:huber}
\end{equation}
where $c$ specifies the tolerance region within which differences between human and model scores are treated as minor (we set c=5 for all experiments). We normalize and average the penalties to obtain a calibration term

\begin{equation}
\psi = \frac{1}{2}\left(\frac{\rho_c(e_A)}{c} + \frac{\rho_c(e_B)}{c}\right),
\end{equation}

\noindent which scales the deviation into approximately $[0,1]$ and increases as the model’s predicted scores deviate from human scores. The final shaped reward is

\begin{equation}
r = r_{\text{rank}} * \big[\,1 - \beta\,\psi\,\big],
\end{equation}

\noindent Here, $c$ sets the tolerance region for small errors and $\beta$ controls the strength of shaping.
We apply shaping only when the ranking is correct ($r_{\text{rank}}=1$), which preserves the simplicity and verifiability of the reward while improving score calibration.
In other words, an incorrect ranking always yields zero reward, even if the predicted scores are numerically close.
The shaping term thus provides score calibration beyond ranking, making the feedback denser while keeping the preference supervision strictly verifiable.
An ablation study on reward shaping is presented in Appendix~\ref{appendix:albation}.
We additionally tested adding an explanation-quality penalty (genRM) as an auxiliary signal, but it yields only a marginal effect.
Therefore, all main results in this paper are reported with the pairwise preference reward plus Huber shaping, without genRM.

\subsection{RLVR Training with PPO}
\label{sec:rlvr-ppo}
We train Remedy-R using reinforcement learning with the verifiable reward in §\ref{sec:reward-design}. The model acts as a policy that, given an input, generates reasoning steps and final scores, and then receives a scalar reward that reflects alignment with human judgments. This objective is to update the model so that high-reward behaviors become more likely over time. Formally, the model defines a conditional policy $\pi_\theta(\mathbf{y}\mid\mathbf{x})$ over output sequences. After generating a response, we compute a scalar reward and maximize the expected return with gradients estimated via the policy gradient theorem:

\begin{equation}
\mathcal{J}(\theta)=\mathbb{E}_{\mathbf{x},\mathbf{y}\sim\pi_\theta}\big[r(\mathbf{y},\mathbf{x})\big], \quad
\nabla_\theta \mathcal{J}(\theta)=\mathbb{E}\Big[\sum_{t=1}^T \nabla_\theta\log\pi_\theta(y_t\mid y_{<t},\mathbf{x})\,A_t\Big].
\end{equation}

\noindent where $A_t$ denotes the token-level advantage. This estimator increases the likelihood of high-reward generations and decreases that of low-reward ones, aligning the model’s behavior with the reward signal. We optimize this objective using Proximal Policy Optimization (PPO)~\citep{schulman2017proximal}, which stabilizes updates by clipping the ratio between the new and the old policy, while regularizing against a frozen reference model $\pi_{\theta_{\mathrm{ref}}}$ to prevent reward over-optimization. During training, each rollout prompts the model with a translation pair, generates an evaluation output, parses the scores, computes the reward, estimates token level advantages, and finally applies PPO updates. We then optimize the following PPO objective:

\begin{equation}
\mathcal{L}_{\text{PPO}}(\theta) =
\mathbb{E}_t \Big[
\min\big(
r_t(\theta) A_t,\ 
\mathrm{clip}(r_t(\theta), 1 - \epsilon, 1 + \epsilon) A_t
\big)
\Big]
 - \beta_{\text{KL}}\,\mathrm{KL}\big[\pi_\theta\,\|\,\pi_{\theta_{\mathrm{ref}}}\big].
\end{equation}

\noindent Here, $r_t(\theta) = \frac{\pi_\theta(y_t \mid y_{<t}, \mathbf{x})}{\pi_{\theta_{\text{old}}}(y_t \mid y_{<t}, \mathbf{x})}$ is the likelihood ratio, $\epsilon$ is the clipping threshold, and $\beta_{\text{KL}}$ controls the strength of the KL penalty that regularizes the updated policy toward a frozen reference policy $\pi_{\theta_{\text{ref}}}$ (the corresponding pretrained base model of the same size). Rewards are terminal and defined on the final parsed score. For advantage estimation, we use Generalized Advantage Estimation (GAE) with $\lambda=1$, which trades off bias and variance by exponentially weighting multi step returns:

\begin{equation}
A_t = \sum_{l=0}^{\infty} (\gamma\lambda)^l \delta_{t+l},
\quad \delta_t = r_t + \gamma V_\phi(s_{t+1}) - V_\phi(s_t).
\end{equation}

\noindent where $V_\phi$ is the value function, and $\gamma\in(0,1]$ and $\lambda\in[0,1]$ are standard discounting hyperparameters. 
Following recent work~\citep{hu2025open}, we set $\lambda=1$, which simplifies the estimator to a discounted Monte Carlo return and improves stability in practice. 

\begin{policybox}[width=\columnwidth]{Remedy-R Training Prompt Template}
\small
You are an expert machine translation evaluator. You need to assess the quality of two translations of the same source text. Your task is to evaluate the translation quality and provide scores from 0--100, where higher scores indicate better quality. You are also given a reference (not always perfect) to help you assess the quality.\\[2pt]
\textbf{Evaluation Criteria:}
\begin{itemize}[leftmargin=*,nosep]
\item Accuracy: Whether the meaning expressed in the translation is correct and faithful to the source. Penalize mistranslation, unsupported additions/hallucinations, terminology errors, and untranslated text.
\item Fluency: How natural and grammatically correct the translation reads in the target language. Consider grammar, agreement, word order, punctuation, spelling, register.
\item Completeness: Is all information from the source conveyed without omissions?
\end{itemize}
\textbf{Instructions:} Think step by step about the quality of each translation and write your analysis first, then provide your final scores. Evaluate each translation independently rather than by comparison.\\[2pt]
\textbf{Output Format:} Thinking through your evaluation first, then output the scores in exactly this format (do not give scores first):\\
\#\#\#\#\\
A: [score]\\
B: [score]\\[2pt]
Now evaluate this \$SRC\_LANG--\$TGT\_LANG translation:\\
\texttt{-----}\\
Source: \$SOURCE\\
Reference: \$REFERENCE\\
Translation A: \$TRANSLATION\_A\\
Translation B: \$TRANSLATION\_B
\end{policybox}

\section{Experimental Setup}\label{sec:Experimental_Setup}

This section describes our training data, evaluation benchmarks, baselines, and implementation details.

\subsection{Training Data}

We train Remedy-R using the WMT20 Multidimensional Quality Metric (MQM) dataset~\citep{mathur2020results}. The MQM dataset is widely regarded as one of the highest-quality MT evaluation resources, as it relies on professional translators, offering higher reliability than crowd-sourced Direct Assessment (DA) data~\citep{freitag2021experts}. The WMT20 MQM data include two language pairs (English–German and Chinese–English). After constructing pairwise preferences from human scores, we obtain approximately 60K training pairs.

\subsection{Meta Evaluation Benchmarks}

We evaluate Remedy-R across two complementary benchmarks that assess both in domain and out of distribution robustness.

\paragraph{WMT Metric Shared Tasks.} 

We use the WMT22–24 MQM benchmarks~\citep{freitag-etal-2022-results, freitag-etal-2023-results, freitag-etal-2024-llms}, which provide standardized suites for meta evaluation. WMT22 includes 54 systems and 106K segments; WMT23 includes 21 systems and 63K segments; and WMT24 (MQM subset) includes 3 language pairs, 32 systems, and 68K segments.

\paragraph{MSLC Challenge Set.} We additionally evaluate on the MSLC~\citep{knowles2024mslc24}，which targets metric robustness under out of distribution and adversarial conditions. It consists of several controlled perturbation types:  
(1) \emph{Empty translation}: the translation is an empty string;  
(2) \emph{Empty source/reference}: the translation is not empty while the source or reference is empty, simulating overgeneration from no input;
(3) \emph{Source copy}: the translation is identical to the source;  
(4) \emph{Wrong language}: valid and fluent translation in an unintended language; and
(5) \emph{Mixed language}: the translation covers the entire source content but mixes multiple languages.
(6) \emph{Unrelated Translation}: the translations are completely unrelated to source texts.
Together, these stress tests assess whether a metric can reliably penalize degenerate, hallucinated, or linguistically inconsistent outputs.

\subsection{Baselines}

    We compare Remedy-R with both open source scalar metrics and closed LLM-as-judge evaluators.
    
    \subsubsection{LLM-as-Judge Metrics}

    \paragraph{GEMBA.} A zero-shot prompting approach using GPT-4~\citep{achiam2023gpt} for translation quality assessment. GEMBA-DA~\citep{kocmi2023large} provides a single quality score; GEMBA-ESA provides error span analysis with a final quality score; GEMBA-MQM~\citep{kocmi2023gemba} provides MQM style annotations that mimic human MQM evaluation.

    \paragraph{PaLM.} Similarly, \citet{fernandes2023devil} evaluate translation quality by prompting PaLM-540B~\citep{chowdhery2023palm} in a zero-shot setting.

    \paragraph{EAPrompt.} An approach that combines chain-of-thought reasoning with error span analysis for LLM-based MT evaluation~\citep{lu2024error}, which improves over GEMBA-DA in both correlation and consistency.

    \paragraph{MQM-APE.} This metric extends GEMBA with an Automatic Post-Editing (APE) stage to filter non-impactful errors~\citep{lu2025mqm}. It consists of three components: (1) an evaluator producing error annotations, (2) a post-editor assessing error impact, and (3) a verifier filtering unimportant errors.

    \subsubsection{Scalar MT Metrics}

    \paragraph{COMET-22.} A widely used neural metric fine-tuned from XLM-R to regress human ratings~\citep{rei2022comet} using extensive WMT metric training data.


    \paragraph{MetricX-XXL.} A state-of-the-art regression-based metric fine-tuned from mT5-XXL (13B)~\citep{juraska2023metricx}, achieving top performance in WMT22–24. We compare to MetricX 22, 23, and 24-hybrid for the corresponding WMT benchmark.

    
    \paragraph{PaLM-2 BISON FT.} An approach that finetunes the second largest BISON model in the PaLM-2 family with regression training objective using WMT metric training data~\citep{fernandes2023devil}.


    \paragraph{Remedy.} A pairwise reward modeling approach that fine-tunes Gemma2-9B via Bradley–Terry optimization, achieving the newest SOTA results on WMT22-24 benchmarks~\citep{tan2025remedy}.

\subsection{Implementation and Meta-Evaluation}

We train Remedy-R on the Qwen2.5 model family across three parameter scales (7B, 14B, and 32B). We use the VeRL framework~\citep{sheng2024hybridflow} for PPO training with FSDP, and employ vLLM~\citep{kwon2023efficient} for efficient rollout inference. All training runs for one epoch with a maximum sequence length of 2048 tokens. We use the Adam optimizer with a learning rate of $5{\times}10^{-6}$ and an effective batch size of 2048. Experiments are run on 4 NVIDIA H100 GPUs for the 7B model and 8 H200 GPUs for the 14B and 32B models. The training takes approximately 10–27 hours to complete.\\

\noindent For meta-evaluation, we adopt the official WMT Metric Shared Task toolkit (MTME).\footnote{\url{https://github.com/google-research/mt-metrics-eval}} Following the official setup, we report Pairwise Accuracy (Acc)~\citep{kocmi2021ship} at the system level and tie-calibrated pairwise accuracy ($\mathit{acc^*_{eq}}$)~\citep{deutsch2023ties} at the segment level, with significance assessed using the Perm-Both test~\citep{deutsch2021statistical}. One exception is WMT24, where we report Soft Pairwise Accuracy (SPA)~\citep{thompson2024improving,freitag-etal-2024-llms} for system-level meta-evaluation.

\section{Automatic Translation Evaluation Results and Analyses}\label{sec:results}
\begin{table}[h!]
\centering
\def\arraystretch{1.0} 
\resizebox{\linewidth}{!}{%
\begin{tabular}{llcccccc||c}
\toprule
\multirow{2}{*}{\textbf{Type}} & 
\multirow{2}{*}{\textbf{Methods}} & 
\multirow{2}{*}{$\boldsymbol{\theta}$} & 
\multicolumn{1}{c}{\textbf{System-Level}} & 
\multicolumn{4}{c}{\textbf{Segment-Level} $\boldsymbol{\mathit{acc^*_{eq}}}$} &
\multicolumn{1}{c}{\textbf{Avg}} \\
\cmidrule(lr){4-4} \cmidrule(lr){5-8}
      & & & \textbf{Acc (3 LPs)} & \textbf{Avg} & \textbf{En-De} & \textbf{En-Ru} & \textbf{Zh-En} & \textbf{Corr} \\
\midrule
\multirow{5}{*}{\makecell[l]{\textbf{Scalar}\\\textbf{Metrics}}} 
 & COMET-22-DA              & 0.5B  & 82.8  & 54.5 & 58.2 & 49.5 & 55.7 & 68.7 \\
 & COMET-22 (ensemble)      & 5x0.5B  & 83.9 & 57.3 & 60.2 & 54.1 & 57.7 & 70.6 \\
 & MetricX-XXL              & 13B   & 85.0  & 58.8 & 61.1 & 54.6 & 60.6 & 71.9 \\
 & PaLM-2 BISON FT             & >100B      & 88.0  & 57.3 & 61.0 & 51.5 & 59.5 & 72.7 \\
 & \textnormal{ReMedy} & 9B  & \colorbox{orange1}{91.2} & \colorbox{orange1}{58.9} & 61.0 & 60.4 & 55.4 & \colorbox{orange1}{75.1} \\

\midrule
\multirow{11}{*}{\makecell[l]{\textbf{LLM}\\\textbf{Judges}}}  
 & EAPrompt (Llama2)               & 70B     & 85.4  & 52.3 & 55.2 & 51.4 & 50.2 & 68.9 \\
 & EAPrompt (Mistral)              & 8x7B     & 84.0  & 50.9 & 53.8 & 50.6 & 48.2 & 67.5 \\
 & EAPrompt (GPT3.5-Turbo)         & >100B       & 91.2  & 53.3 & 56.7 & 53.3 & 50.0 & 72.3 \\
 & GEMBA-MQM (Qwen)                & 72B     &  84.7  & 53.8 & 56.0 & 54.7 & 50.6 & 69.3 \\
 & MQM-APE (Qwen)                  & 72B     & 85.8  & 54.5 & 56.4 & 55.7 & 51.4 & 70.2 \\
 & MQM-APE (Mistral)               & 8x22B   & 88.3  & 54.2 & 56.9 & 55.1 & 50.6 & 71.3 \\
 & GEMBA-DA (GPT4)                    & >100B      & 89.8  & 55.6 & 58.2 & 55.0 & 53.4 & 72.7 \\
 & PaLM                            & 540B    & 90.1  & 50.8 & 55.4 & 48.6 & 48.5 & 70.5 \\
 
\cmidrule(lr){2-9}
& \textnormal{Remedy-R}  & 7B  & 89.1 & 54.8 & 58.0 & 56.0 & 50.4 & 71.9 \\

& \textnormal{Remedy-R}  & 14B  & 88.7 & \colorbox{green2}{56.0} & 58.0 & 55.8 & 54.2 & 72.4 \\

& \textnormal{Remedy-R}  & 32B  & \colorbox{green2}{91.6} & 55.2 & 57.8 & 55.7 & 52.2 & \colorbox{green2}{73.4} \\

\bottomrule 
\end{tabular}%
}
\caption{Evaluation on WMT22 MQM set. Following official WMT22 settings, we report system-level Pairwise Accuracy (Acc) and segment-level pairwise accuracy with tie calibration ($\mathit{acc^*_{eq}}$), using Perm-Both statistical significance test~\citep{deutsch2021statistical}. \colorbox{orange1}{orange} and \colorbox{green2}{blue} indicate the best performing \colorbox{orange1}{scalar} and \colorbox{green2}{LLM judge} metrics. Our Remedy-R 7B outperforms all LLM judges with 70B parameters, and Remedy-R 32B surpasses GEMBA-DA (GPT4).}
\label{tab:wmt22}
\end{table}

\noindent In this section, we evaluate Remedy-R on automatic translation evaluation tasks and address three questions: (a) how Remedy-R compares with top scalar MT metrics and LLM-as-judge (§\ref{sec:wmt_bench}); (b) how performance changes under test-time-scaling with multiple evaluation passes (§\ref{sec:tts}); and (c) whether the model behaves robustly on out-of-domain stress tests (§\ref{sec:ood}).

\subsection{Meta Evaluation on WMT benchmarks}\label{sec:wmt_bench}


\noindent As shown in Table~\ref{tab:wmt22}, Remedy-R achieves leading performance on the WMT22 metric benchmark compared to both commercial LLM judges and open metrics. Specifically, Remedy-R 7B surpasses all open LLM-as-judge baselines with 70B parameters (e.g., EAPrompt and MQM-APE), while Remedy-R 32B exceeds GEMBA-DA (GPT-4). Among scalar metrics, Remedy-R also outperforms the regression-based PaLM-2 BISON and MetricX-XXL, showing that our model aligns with human ratings on human translation quality judgments.\\

\noindent In addition, we also provide additional results on WMT23 and WMT24 metric benchmarks in Table~\ref{tab:wmt23} and Table~\ref{tab:wmt24}. The results highlight that Remedy-R achieves on-par or superior performance to the current SOTA metrics. For example, on wmt23, Remedy-R-7B outperforms EAPrompt that is based on GPT4o-mini, and Remedy-R 14B and 32B performs on par with KIWI-XXL and MetricX-23. On wmt24, Remedy-R outperforms GEMBA-ESA (GPT4), MetricX-24-Hybrid, and XCOMET-XXL (see details in Table~\ref{tab:wmt24}).

\subsection{Test Time Scaling with multiple Evaluation Passes}\label{sec:tts}

Remedy-R’s generative reasoning nature enables the application of \textbf{Test-Time Scaling (TTS)}, where multiple evaluation passes are performed with different reasoning trajectories and their quality scores are aggregated.
In this setting, we adopt a simple implementation that averages the quality scores from multiple independent evaluations.

\begin{figure*}[h!]
    \centering
    \includegraphics[width=0.8\linewidth]{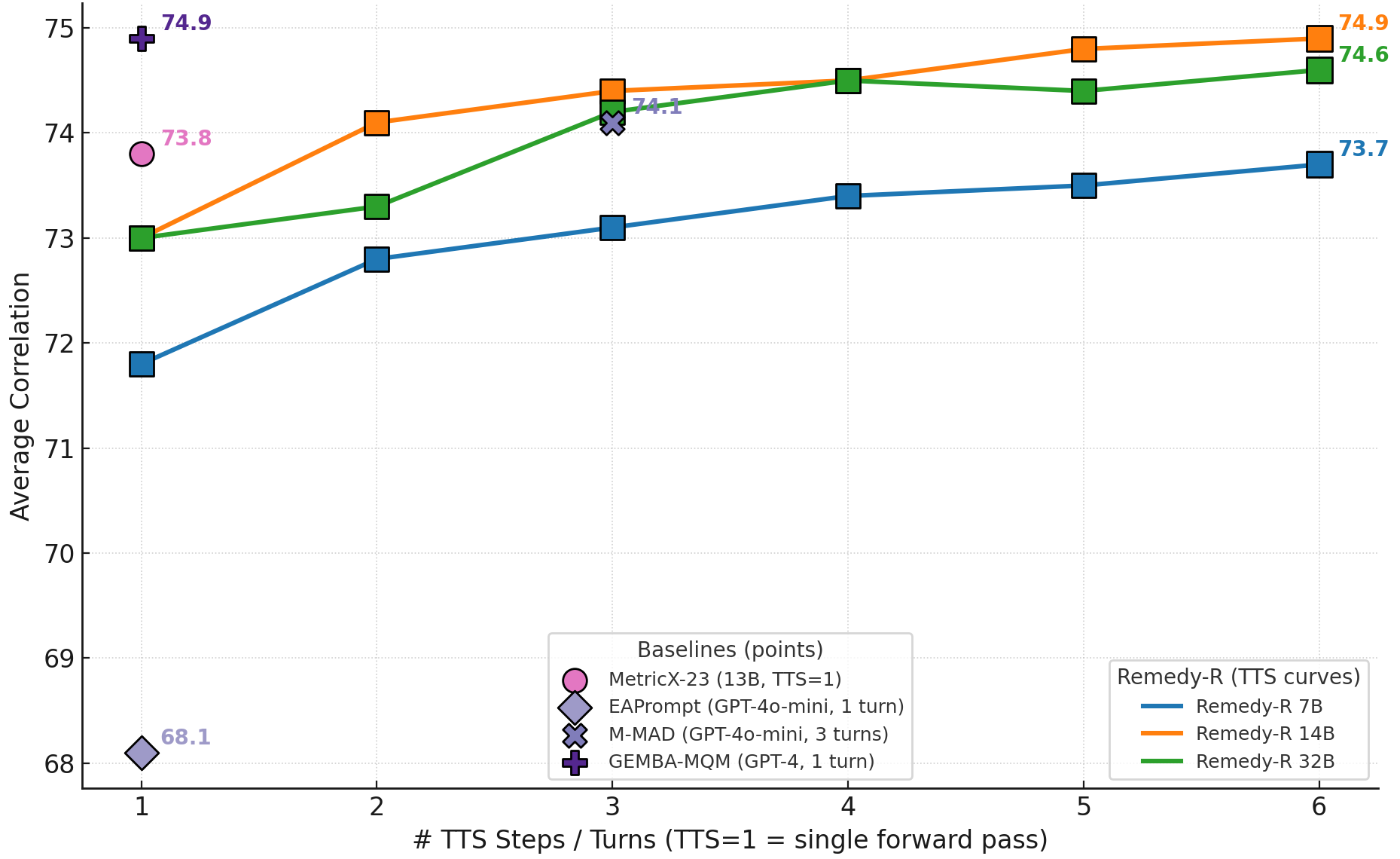}
    \caption{ 
    Average correlation across WMT23 MQM benchmarks under different numbers of Test-Time Scaling (TTS) evaluation passes. Each configuration aggregates multiple independent evaluations by averaging their final quality scores. TTS consistently improves correlation as the number of evaluation passes increases. Full results are shown in Table~\ref{tab:wmt23} in Appendix.}
    \label{fig:tts}
\end{figure*}

\noindent As shown in Figure~\ref{fig:tts}, performing more evaluation trajectories at test time consistently enhances performance across all model sizes. Notably, Remedy-R-14B reaches an average correlation of 74.9, matching the strongest GEMBA-MQM performance.
The steady improvement from 7B to 32B suggests that iterative reasoning stabilizes evaluation outcomes and reduces stochastic variance, yielding more robust and reliable quality assessments.\\

\noindent Interestingly, we observe that TTS primarily improves \textit{segment-level} $\mathit{acc^*_{eq}}$ rather than system-level correlation (see Table~\ref{tab:wmt23}). We hypothesize that this phenomenon is mostly due to the limitations of current meta-evaluation metrics. As noted by \citet{perrella2024guardians}, tie-calibrated pairwise accuracy ($\mathit{acc^*_{eq}}$) tends to favor metrics that output continuous rather than discrete scores. Averaging multiple predictions effectively smooths discrete outputs into continuous scores, improving agreement with tie-calibrated accuracy, which favors metrics with finer score granularity.

\subsection{Analyses on Challenge sets}\label{sec:ood}

We further evaluate Remedy-R under \textit{out-of-distribution} (OOD) and adversarial conditions using the MSLC24 challenge set. To extend coverage, we additionally sample 50 random target-language sentences from Flores-200~\citep{costa2022no} to construct an extra unrelated-translation category. For most perturbation types, such as empty outputs, source copies, wrong-language translations, and unrelated MTs, a reliable metric should assign quality scores as low as possible. For the mixed-language setting, where translations remain semantically correct but include code-switching, are expected to receive moderately higher scores rather than near-zero ones.\\

\begin{table*}[h!]
\centering
\def\arraystretch{1.0}%
\resizebox{0.7\linewidth}{!}{%
\begin{tabular}{lccccccccccc}
\toprule

& \multirow{2}{*}{\textbf{ref?}}  & \textbf{empty} & \textbf{empty} & \textbf{src} & \textbf{wrong} & \textbf{mix} & \textbf{unrelated} \\
& & \textbf{mt} & \textbf{src+ref} & \textbf{copy} & \textbf{lang} & \textbf{lang} & \textbf{mt}\\
\midrule
COMET-22 & \cmark & 57.00\% & 58.81\% & 69.85\% & 67.84\% & 65.56\% & 45.23\% \\
KIWI & \xmark & 54.87\% & 67.72\% & 52.15\% & 82.64\% & 78.75\% & 41.95\% \\
XCOMET & \cmark & 73.79\% & 64.12\% & 82.04\% & 85.65\% & 71.77\% & 20.31\% \\
MetricX-24-XXL & \cmark & -9.59 & -5.85 & -12.59 & -3.06 & -10.08 & -24.15 \\
MetricX-24-XXL & \xmark & -7.34 & -5.85 & -11.36 & -2.51 & -7.78 & -24.25 \\
GEMBA-ESA & \xmark & 14.00\% & 13.5\% & 11.12\% & 14.32\% & 18.08\% & 1.27\% \\
\midrule
ReMedy-R-7B  & \cmark & 1.00\% & 7.07\% & 76.92\% & 43.69\% & 60.6\% & 1.5\% \\
ReMedy-R-7B  & \xmark & 0.83\% & 5.40\% & 90.38\% & 33.15\% & 65.4\% & 2.0\% \\
ReMedy-R-14B & \cmark & 0.00\% & 0.00\% & 11.35\% & 14.6\% & 37.6\% & 0.6\% \\
ReMedy-R-14B & \xmark & 0.00\% & 0.00\% & 28.07\% & 12.1\% & 35.5\% & 1.0\% \\
ReMedy-R-32B & \cmark & 0.00\% & 0.00\% & 2.76\% & 8.30\% & 46.0\% & 1.3\% \\
ReMedy-R-32B & \xmark & 0.00\% & 0.00\% & 1.30\% & 8.0\% & 43.8\% & 3.5\% \\

\bottomrule
\end{tabular}%
}
\caption{Averaged quality scores of different metric models on MSLC24 OOD set. For all classes except \textit{mix-lang}, a robust metric should output low scores; for \textit{mix-lang}, the translation preserves the source meaning but contains code-switching, so its quality scores should be moderately high rather than near zero. MetricX scores are ranged from -25 to 0.}
\label{tab:MSLC}
\end{table*}

\noindent As shown in Table~\ref{tab:MSLC}, scalar MT metrics (e.g., COMET-22 and MetricX-24) often fail to produce sensible scores under these OOD conditions, sometimes even assigning high quality to nonsensical or copy outputs. For instance, XCOMET produces 73.8\%, 64.1\%, and 82.0\% for the empty translation, empty source and reference, and source copy settings, respectively. In contrast, Remedy-R exhibits strong robustness and controlled sensitivity: it consistently outputs near-zero scores for empty translations, sharply penalizes source-copy and wrong-language cases, and yields moderate scores for mixed-language inputs.\\

\noindent Notably, unlike MetricX-24 and XCOMET, which are trained with mixed synthetic data covering empty or hallucinated translations, Remedy-R is trained without synthetic augmentation. This further highlights that Remedy-R maintains coherent behavior across diverse OOD perturbations, effectively distinguishing between semantic degradation and benign linguistic variation. Such robustness suggests that its evaluation judgments generalize beyond standard WMT data and remain reliable even under adversarial translation conditions.

\section{ReMedy-R Agent}\label{sec:agent}

So far, our analyses have focused on evaluation at the final score level, that is, how well Remedy-R’s predicted quality scores align with human ratings.  
However, such correlations do not necessarily imply that the model’s intermediate \textit{reasoning explanations} are faithful or useful. To examine whether Remedy-R’s explanations truly reflect actionable translation quality judgments, we take two complementary steps:
(1) we prompt GPT-4o-mini to directly score the faithfulness of Remedy-R’s explanations given only the source sentence and the translation hypothesis (§\ref{sec:agent-faithfulness}); and (2) we go one step further by treating the explanations as feedback in a simple \textbf{evaluate–revise} probe (\textbf{Remedy-R Agent}) and measuring their impact on translation quality (§\ref{sec:agent-1}-\ref{sec:agent-3}).\\

\noindent For the evaluate–revise Agent setting (§\ref{sec:agent-1}-\ref{sec:agent-3}), three models are involved:
\begin{itemize}[leftmargin=1.2em,itemsep=2pt]
    \item \(M_{\textit{base}}\): the initial translation model that produces initial translation given a source sentence.
    \item \(M_{\textit{feedback}}\): the evaluation model that inspects the initial translation and generates a reasoning-based quality assessment (e.g.: accuracy, fluency, completeness) with an explicit explanation.
    \item \(M_{\textit{refinement}}\): the refinement model that revises the translation based on \(\{\,\texttt{src},\,\texttt{mt},\,\texttt{feedback}\,\}\).
\end{itemize}

\noindent When all three roles are instantiated, we obtain a full \textbf{Remedy-R Agent} that not only evaluates translations but also uses its own chain-of-thought evaluation to guide refinement.  By observing how much improvement the feedback induces, we can assess both the \emph{faithfulness} and the \emph{utility} of Remedy-R’s explanations in practice.\\

\noindent Specifically, we design a series of controlled experiments to evaluate this agent framework under different configurations of \(M_{\textit{base}}\), \(M_{\textit{feedback}}\), and \(M_{\textit{refinement}}\).  
These experiments allow us to answer three key questions:  
(i) Are Remedy-R’s explanations genuinely \textit{faithful and useful} for improving translations?
(ii) Can Remedy-R function as a multi-task evaluator–refiner?
(iii) Does Remedy-R Agent Remain Effective on SOTA Translation Systems?
We describe each setup in detail below (see Table~\ref{tab:Agent-qwen} and Figure~\ref{fig:refine}).

\subsection{How faithful are the evaluation explanations?}\label{sec:agent-faithfulness}

\begin{table}[h!]
\centering
\def\arraystretch{1.0}%
\setlength{\tabcolsep}{6pt}
\begin{tabular}{lcccc}
\toprule
\textbf{Remedy-R} & \textbf{en-de} & \textbf{en-ru} & \textbf{zh-en} & \textbf{Avg} \\
\midrule
7B  & 79.10 & 76.57 & 74.95 & 76.87 \\
14B & 80.02 & 76.75 & 78.05 & 78.27 \\
32B & 81.18 & 79.05 & 78.38 & 79.54 \\
\bottomrule
\end{tabular}
\caption{GPT-4o-mini faithfulness scores for Remedy-R explanations on WMT22 MQM test samples (300 per language pair). GPT-4o-mini is given the source, translation, and explanation. Higher score means more faithful explanations.}
\label{tab:gpt4-faithfulness}
\end{table}

\noindent We first assess explanation faithfulness by prompting GPT-4o-mini to score how well Remedy-R’s reasoning is supported by the source sentence and the translation hypothesis only. We conduct this analysis on the WMT22 metric evaluation test set and sample 300 examples for each MQM language pair (en-de, en-ru, zh-en), resulting in 900 examples in total.
Table~\ref{tab:gpt4-faithfulness} reports the average faithfulness scores. Overall, Remedy-R explanations receive consistently high faithfulness scores, and faithfulness improves with model scale.

\subsection{How useful are the evaluation explanations for refinement?}\label{sec:agent-1}

We first examine whether evaluation explanations provide useful guidance beyond self-refinement.
To this end, we fix the refinement model to the same translator that produced the initial translation (\(M_{\textit{refinement}} = \textit{Base}\)) and compare two setups:
(1) self-refinement (rows where \(M_{\textit{feedback}} = -\), \(M_{\textit{refinement}} = \textit{Base}\)), where the base model revises its own translation without feedback, and
(2) feedback-refinement(rows where \(M_{\textit{feedback}} = \textit{Remedy-R}\), \(M_{\textit{refinement}} = \textit{Base}\)), where the base model receives Remedy-R’s evaluation reasoning analysis as additional context.
Across all model sizes, the feedback-refinement setup yields consistent quality improvements, confirming that Remedy-R’s explanations contain faithful information that helps guide effective refinements.\\

\noindent We further compare Remedy-R with \textbf{x-Tower}~\citep{treviso2024xtower}, a GPT-4–distilled LLMs that produces descriptive explanations but relies on external span annotations from \textbf{xCOMET-XL}~\citep{guerreiro2024xcomet} for error identification. Unlike x-Tower, Remedy-R directly outputs both reasoning and scalar scores without any external supervision. As shown in Table~\ref{tab:Agent-qwen} (rows where \(M_{\textit{refinement}} = \textit{Base}\)), Remedy-R feedback achieves larger and more stable improvements than x-Tower at similar model scales, despite not using distillation or fine-grained annotations.
Even at 7B, its feedback remains competitive against the combined x-Tower (14B + 3.5B with xCOMET), highlighting that Remedy-R’s reasoning explanations are genuinely useful for driving translation refinement.

\begin{table*}[h!]
\centering
\def\arraystretch{1.0}%
\setlength{\tabcolsep}{0.5pt}
\resizebox{\linewidth}{!}{%
\begin{tabular}{llccccccccccc!{\hspace{2pt}\vrule width 1.2pt\hspace{2pt}}c}
\toprule
\(M_{\textit{feedback}}\) & \(M_{\textit{refinement}}\) & \textbf{cs-uk} & \textbf{en-cs} & \textbf{en-de} & \textbf{en-es} & \textbf{en-hi} & \textbf{en-is} & \textbf{en-ja} & \textbf{en-ru} & \textbf{en-uk} & \textbf{en-zh} & \textbf{ja-zh} & \textbf{Avg} \\

\midrule
\multicolumn{14}{c}{\textbf{\textit{\(M_{\textit{base}}\) = Qwen2.5-it-7B | Remedy-R = 7B | x-Tower = 14B (w. XComet-XL = 3.5B)}}} \\
\rowcolor[gray]{0.94}
 - & - & 62.9 & 53.1 & 86.0 & 81.9 & 37.9 & 29.8 & 69.7 & 71.2 & 51.5 & 82.8 & 69.9 & 63.4 \\

 - & Base & 64.3\chg{+1.4} & 55.0\chg{+1.9} & 86.4\chg{+0.4} & 82.1\chg{+0.2} & 39.1\chg{+1.2} & 30.7\chg{+0.9} & 69.6\chg{-0.1} & 72.1\chg{+0.9} & 53.8\chg{+2.3} & 82.7\chg{-0.1} & 70.4\chg{+0.5} & 64.2\chg{+0.8} \\
 
x-Tower & x-Tower           & 75.6\chg{+12.6} & 57.0\chg{+3.9} & 90.7\chg{+4.7} & 85.8\chg{+3.9} & 40.3\chg{+2.4} & 33.6\chg{+3.7} & 62.3\chg{-7.4} & 79.3\chg{+8.0} & 69.2\chg{+17.7}  & 82.2\chg{-0.6} & 66.5\chg{-3.4} & 67.5\chg{+4.2} \\

Remedy-R \hspace{2mm} & Remedy-R \hspace{2mm} & 
66.0\chg{+3.1} & 56.0\chg{+2.9} & 87.2\chg{+1.1} & 83.2\chg{+1.3} & 40.5\chg{+2.6} & 30.9\chg{+1.1} & 71.8\chg{+2.1} & 73.4\chg{+2.1} & 54.6\chg{+3.1} & 83.2\chg{+0.5} & 69.7\chg{-0.2} & 65.1\chg{+1.8} \\

x-Tower & Base   & 66.3\chg{+3.4}  & 55.8\chg{+2.6} & 87.3\chg{+1.3} & 83.0\chg{+1.1} & 39.4\chg{+1.5} & 30.3\chg{+0.5} & 70.4\chg{+0.7} & 74.2\chg{+2.9} & 56.0\chg{+4.4 }  & 83.0\chg{+0.2} & 69.7\chg{-0.2} & 65.0\chg{+1.7} \\
Remedy-R & Base \hspace{2mm} & 65.8\chg{+2.9}  & 55.9\chg{+2.8} & 86.9\chg{+0.9} & 82.9\chg{+1.0} & 40.4\chg{+2.5} & 30.5\chg{+0.7} & 71.5\chg{+1.8} & 73.3\chg{+2.1} & 54.4\chg{+2.9 }  & 83.0\chg{+0.3} & 69.3\chg{-0.6} & 64.9\chg{+1.6} \\

\midrule
\multicolumn{14}{c}{\textbf{\textit{\(M_{\textit{base}}\) = Qwen2.5-it-14B | Remedy-R = 14B | x-Tower = 14B (w. XComet-XL = 3.5B)}}} \\
\rowcolor[gray]{0.94}
- & -                & 69.4 & 63.6 & 88.4 & 83.7 & 47.6 & 32.6 & 74.8 & 75.7 & 58.6 & 83.9 & 72.5 & 68.2 \\

- & Base & 71.0\chg{+0.6} & 67.2\chg{+3.6} & 89.4\chg{+1.0} & 85.0\chg{+1.3} & 51.5\chg{+3.9} & 34.4\chg{+1.9} & 77.9\chg{+3.1} & 77.9\chg{+2.2} & 63.2\chg{+4.6} & 84.5\chg{+0.6} & 72.7\chg{+0.2} & 70.4\chg{+2.2}\\

x-Tower & x-Tower & 77.1\chg{+7.7} & 62.5\chg{-1.1} & 91.4\chg{+3.0} & 86.5\chg{+2.8} & 45.0\chg{-2.6} & 34.1\chg{+1.6} & 65.6\chg{-9.2} & 80.2\chg{+4.5} & 70.4\chg{+11.8}  & 82.8\chg{-1.1} & 68.3\chg{-4.2} & 69.4\chg{+1.2} \\
Remedy-R & Remedy-R & 74.1\chg{+4.7} & 68.3\chg{+4.7} & 89.8\chg{+1.4} & 84.9\chg{+1.2} & 52.6\chg{+5.0} & 35.8\chg{+3.2} & 77.8\chg{+2.9} & 77.4\chg{+1.7} & 64.3\chg{+5.7 }  & 83.8\chg{-0.1} & 72.4\chg{-0.1} & 71.0\chg{+2.8} \\
x-Tower & Base   & 73.0\chg{+3.6} & 68.5\chg{+4.9} & 90.2\chg{+1.8} & 85.3\chg{+1.7} & 51.3\chg{+3.7} & 34.6\chg{+2.0} & 78.6\chg{+3.8} & 78.5\chg{+2.8} & 65.1\chg{+6.5 }  & 84.3\chg{+0.4} & 71.9\chg{-0.6} & 71.0\chg{+2.8} \\
Remedy-R & Base  & 74.8\chg{+5.4} & 68.2\chg{+4.6} & 90.3\chg{+1.9} & 85.6\chg{+1.9} & 53.1\chg{+5.5} & 36.3\chg{+3.7} & 79.3\chg{+4.5} & 79.2\chg{+3.5} & 64.5\chg{+5.9 }  & 84.1\chg{+0.2} & 72.4\chg{-0.1} & 71.6\chg{+3.4} \\
\midrule
\multicolumn{14}{c}{\textbf{\textit{\(M_{\textit{base}}\) = Qwen2.5-it-32B | Remedy-R = 32B | x-Tower = 14B (w. XComet-XL = 3.5B)}}} \\

\rowcolor[gray]{0.94}
- & -                  & 74.0           & 68.2           & 89.6           & 84.6           & 53.4           & 35.1           & 78.0             & 76.9           & 64.0           & 83.5           & 72.6           & 70.9 \\

- & Base & 75.0\chg{+1.0} & 69.5\chg{+1.3} & 90.0\chg{+0.4} & 85.5\chg{+0.9} & 55.4\chg{+2.0} & 36.5\chg{+1.4} & 78.7\chg{+0.7} & 77.2\chg{+0.3} & 66.7\chg{+2.7} & 83.8\chg{+0.3} & 73.6\chg{+1.0} & 72.0\chg{+1.1}\\

x-Tower & x-Tower           & 77.5\chg{+3.5} & 63.5\chg{-4.7} & 91.7\chg{+2.1} & 86.1\chg{+1.5} & 45.5\chg{-7.9} & 34.2\chg{-0.9} & 66.7\chg{-11.2}  & 80.2\chg{+3.2} & 71.6\chg{+7.6} & 82.5\chg{-1.0} & 68.2\chg{-4.5} & 69.8\chg{-1.1} \\
Remedy-R & Remedy-R         & 76.6\chg{+2.6} & 71.5\chg{+3.3} & 91.1\chg{+1.5} & 85.6\chg{+1.0} & 57.1\chg{+3.7} & 37.4\chg{+2.2} & 80.4\chg{+2.5 }  & 78.5\chg{+1.6} & 67.8\chg{+3.8} & 84.0\chg{+0.6} & 73.3\chg{+0.7} & 73.0\chg{+2.1} \\
x-Tower & Base   & 75.6\chg{+1.6} & 70.4\chg{+2.2} & 91.2\chg{+1.6} & 86.1\chg{+1.5} & 56.1\chg{+2.7} & 36.5\chg{+1.4} & 80.5\chg{+2.6 }  & 80.3\chg{+3.3} & 69.1\chg{+5.1} & 84.8\chg{+1.3} & 72.7\chg{+0.1} & 73.0\chg{+2.1} \\
Remedy-R & Base  & 76.9\chg{+2.9} & 71.5\chg{+3.3} & 91.0\chg{+1.4} & 85.9\chg{+1.3} & 56.8\chg{+3.4} & 37.2\chg{+2.1} & 81.1\chg{+3.2 }  & 79.4\chg{+2.5} & 67.8\chg{+3.8} & 83.9\chg{+0.5} & 73.4\chg{+0.8} & 73.2\chg{+2.3} \\

\bottomrule
\end{tabular}%
}
\caption{Agent MT experiments on WMT24 benchmark using Qwen2.5 series models as the initial \textbf{\textit{\(M_{\textit{base}}\)}} translators (gray background). We report XCOMET-XXL in this table.}
\label{tab:Agent-qwen}
\end{table*}

\subsection{Can Remedy-R Act as Both Evaluator and Refiner?}\label{sec:agent-2}

We next test whether Remedy-R can serve as both the evaluator and the refiner, i.e., \(M_{\textit{feedback}} = M_{\textit{refinement}} = \text{Remedy-R}\). In this unified setup, the model first performs evaluation reasoning and then refines the translation based on its own analysis—without any external feedback or editing supervision.
This setting directly probes whether Remedy-R’s reasoning is coherent enough to drive self-improvement.\\

\noindent We first compare the unified Remedy-R agent with a variant where only the feedback originates from Remedy-R while the refinement is done by the base translator (\(M_{\textit{feedback}} = \text{Remedy-R}, M_{\textit{refinement}} = M_{\textit{base}}\)). As shown in Table~\ref{tab:Agent-qwen}, both configurations achieve similar performance, indicating that Remedy-R’s reasoning is well aligned with its own generation dynamics and can be effectively reused for self-refinement.\\

\noindent Compared to the \textbf{x-Tower} agent~\citep{treviso2024xtower}, which relies on GPT-4 distillation and span-level annotations from \textbf{xCOMET-XL}~\citep{guerreiro2024xcomet}, Remedy-R achieves comparable or higher gains across most language pairs and scales.
When both are of similar size (e.g., Qwen2.5-14B), x-Tower shows regressions on several directions (en-hi, en-ja, en-uk), while Remedy-R maintains steady improvements.
This consistency underscores Remedy-R’s robustness: despite never being trained for editing, its reasoning remains grounded and stable enough to guide its own refinement.

\subsection{Does Remedy-R Agent Remain Effective on SOTA Translation Systems?}\label{sec:agent-3}

\begin{figure*}[h!]
    \centering
    \includegraphics[width=\linewidth]{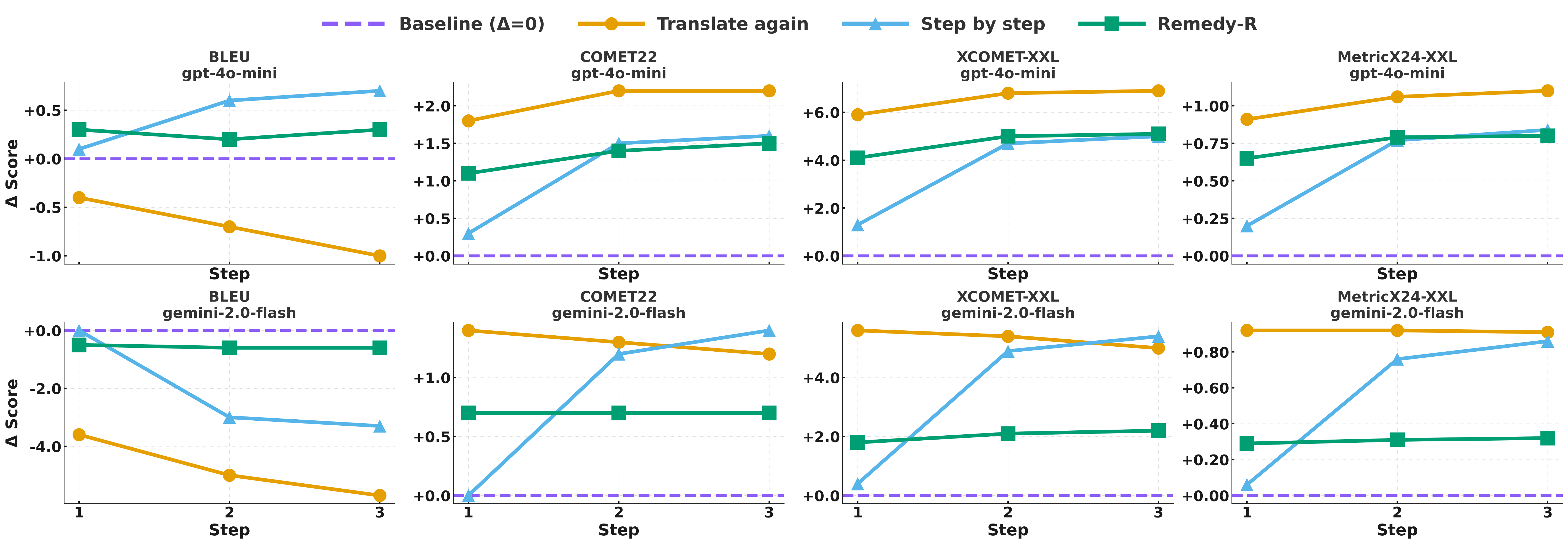}
    \caption{Refinement performance comparison on the initial translations from GPT-4o-mini and Gemini-2.0-Flash using paragraph-level WMT24++ benchmark. We use reference-based XCOMET-XXL measure the translation quality, and provide more metric results in Appendix.}
    \label{fig:refine}
\end{figure*}

\noindent Finally, we evaluate whether Remedy-R Agent remains effective when applied to SOTA translation systems. We set \(M_{\textit{base}}\) to two categories of strong translators: (a) the open-source ALMA-R models~\citep{xu2024paradigm}, built on LLaMA2 and achieving top performance in recent WMT benchmarks, and (b) powerful SOTA commercial LLMs: GPT-4o-mini and Gemini-2.0-Flash. Aligning with~\cite{briakou2024translating, wu2025please}, all experiments are conducted on the WMT24++ translation benchmark~\citep{kocmi2024findings}. To avoid metric interference~\citep{pombal2025adding}, we report results using multiple reference-based metrics, including chrF++, XCOMET-XXL, and MetricX24-XXL.\\

\noindent For the commercial LLMs, we compare Remedy-R Agent with their internal self-refinement strategies: (1) step-by-step~\citep{briakou2024translating}, a multi-step translation pipeline that achieved SOTA results on WMT24, and (2) translate-again~\citep{wu2025please}, which simply regenerates translations for iterative refinement. While these self-refinement strategies rely on the full capacity of large commercial models, our Remedy-R Agent uses only a 32B model.\\

\noindent For open-sourced ALMA-R models, Remedy-R Agent improves translations across multiple language pairs, demonstrating its ability to enhance translations from architectures beyond its own pretraining family (see Table~\ref{tab:alma-r-double}). In addition, Figure~\ref{fig:refine} shows Remedy-R Agent consistently improves translation quality even when starting from strong commercial translations. In specific, on GPT-4o-mini, it is comparable to the step-by-step approach, while on Gemini-2.0-Flash it achieves roughly half of the gain obtained by self-refinement. Given that Remedy-R-32B is significantly smaller than these closed LLMs, these results underscore its robustness and transferability as a lightweight evaluate-refine agent model that generalizes across model families.

\section{Related Work}

\subsubsection{Automatic Translation Evaluation}

Early automatic metrics for machine translation (MT), such as BLEU~\citep{papineni2002bleu} and ChrF~\citep{popovic2015chrf}, rely on surface-level matching between system outputs and references.
Although widely adopted, these metrics show weak correlation with human preferences~\citep{freitag-etal-2022-results}. In contrast, learned neural scalar metrics~\citep{rei-etal-2020-comet,juraska2023metricx,tan2025remedy} achieve much higher and sometimes even superhuman agreement with human ratings on WMT shared tasks~\citep{proietti2025has}. However, they remain black boxes with limited explainability and often fail under out-of-distribution (OOD) conditions~\citep{lo2023metric,knowles2024mslc24,amrhein-etal-2022-aces}.\\

\noindent To enhance explainability, a line of work explores error-based supervision for MT evaluation.
xCOMET~\citep{guerreiro2024xcomet} predicts token-level error spans and their severity from fine-grained MQM annotations, offering localized feedback on translation errors. Building upon this, xTower~\citep{treviso2024xtower} constructs supervised training data by combining xCOMET’s span predictions with GPT-4-generated explanations, and fine-tunes a multi-criteria evaluator on this synthetic feedback.
While this approach improves interpretability under the MQM framework, it heavily depends on explicit span supervision and GPT-4 data, and its generalization to unseen languages and domains remains limited.\\

\noindent Recently, LLM-as-judge approaches (e.g., GEMBA-MQM~\citep{kocmi2023gemba}, EAPrompt~\citep{lu2024error}, MQM-APE~\citep{lu2025mqm}) prompts LLMs for translation evaluation. These systems heavily rely on commericial LLMs like GPT-4, and when switching to open LLMs, their performance remain largely behind. Furthermore, \cite{qian2024large} shows that chain-of-thought prompting provides limited gains for MT evaluation~\citep{qian2024large}. This motivates the need for open metric that integrate decision-making with explicit reasoning, as pursued in Remedy-R.

\subsubsection{Reasoning for Machine Translation}

Recent research has increasingly explored reasoning as a paradigm for both machine translation (MT) evaluation and generation. For MT evaluation, recent LLM-as-judge methods encourage reasoning via multi-stage~\citep{feng2025mad} or multi-agent~\citep{zhang2025himate} frameworks to improve MT evaluation. For MT generation, several recent work explores using reinforcement learning to improve translation quality. For example, MT-R1-Zero~\citep{feng2025mt} adapts the R1~\citep{guo2025deepseek} learning paradigm to MT using Comet and BLEU as reward models, while Hunyuan-MT~\citep{zheng2025hunyuan} mixes XCOMET and DeepSeek-V3~\citep{deepseek2024deepseek} as reward models. In contrast, Remedy-R focus on leveraging efficient deterministic verifiable reward function for MT evaluation.

\subsubsection{Translation Agents}

A growing body of work explores agentic translation, where models iteratively improve their outputs. Recent studies such as translate-again~\citep{wu2025please} and step-by-step~\citep{briakou2024translating} explore self-refinement and translation-subtask paradigms. Other frameworks extend existing evaluators for post-editing: xTower~\citep{treviso2024xtower} uses error-span predictions to produce explanations and corrections. Unlike these studies, our work does not aim to build a new agentic MT pipeline. We adopt the evaluate–revise loop solely as a lightweight probe to test whether the evaluator’s reasoning is faithful, rather than as a new agent paradigm.


\section{Conclusions}


A longstanding challenge in machine translation evaluation is the lack of explainability, which limits the reliability and robustness of current metrics.
In this work, we take a step toward addressing this issue by introducing Remedy-R, a reasoning generative MT metric trained with reinforcement learning via deterministic verifiable rewards built from pairwise human preferences without error annotations. Remedy-R explicitly analyzes translation quality with reasoning COTs then provides a final quality score. Remedy-R presents competitive performance on WMT22–24 metric benchmarks, while remaining robust under OOD stress tests. Beyond correlations, we further verified the faithfulness and utility of its reasoning COTs by using it as feedback in an evaluate–revise loop: Remedy-R Agent consistently improves translations across model families and language pairs, including strong open and even commercial systems.





\bibliography{anthology,custom}

\appendix

\section{Appendix}\label{sec:appendix}

\subsection{Ablation study on WMT22 Metric Benchmark}\label{appendix:albation}

\noindent We conduct an ablation study on the WMT22 metric benchmark to isolate the effect of our reward design during RLVR training.
Unless otherwise specified, all ablations in this subsection are performed with \textbf{Remedy-R 7B}.
We compare two reward settings: (1) a pure pairwise ranking reward that only verifies whether the model-implied preference matches the human preference label; and (2) adding Huber reward shaping to encourage better calibration and reduce overly discrete score behaviors.\\

\noindent As shown in Table~\ref{tab:ablations}, adding Huber reward shaping consistently improves accuracy at both the system and segment levels.
In particular, Huber shaping improves system-level accuracy from 87.6\% to 89.1\% and segment-level accuracy from 52.7\% to 54.8\%, yielding a +1.7\% absolute gain in average accuracy.
These results suggest that Huber shaping provides a more informative learning signal beyond binary preference verification and leads to better-aligned quality judgments.\\

\begin{table}[h!]
\centering
\def\arraystretch{1.08}%
\setlength{\tabcolsep}{4pt}
\resizebox{0.72\linewidth}{!}{%
\begin{tabular}{llccc}
\toprule
\textbf{Model} & \textbf{Reward Setting} & \textbf{Sys} & \textbf{Seg} & \textbf{Avg} \\
\midrule
\multirow{2}{*}{Remedy-R 7B}
& Pairwise Ranking Reward            & 87.6\% & 52.7\% & 70.2\% \\
& + Huber Reward Shaping             & 89.1\% & 54.8\% & 71.9\% \\
\midrule
\multirow{2}{*}{Remedy-R 14B}
& Pairwise + Huber                   & 88.7\% & 56.0\% & 72.4\% \\
& Pairwise + Huber + genRM penalty   & 89.8\% & 56.3\% & 73.1\% \\
\bottomrule
\end{tabular}%
}
\caption{Ablation study on WMT22 comparing reward designs. We report accuracy at the system and segment levels, and their average.}
\label{tab:ablations}
\end{table}

\noindent We further explore incorporating an explanation quality penalty into the reward.
Instead of training a separate reward model, we reuse the same base instruct model (the RL initialization) as a generative rationale judge (genRM). Given the source, translation, and the model-generated explanation, genRM produces a scalar score intended to reflect the rationale's faithfulness and relevance.
We subtract this score (scaled by a fixed coefficient) from the RL reward, penalizing low-quality explanations while keeping the pairwise preference verification term unchanged. As shown in Table~\ref{tab:ablations}, adding this penalty results in only a marginal change in accuracy in our Remedy-R 14B setting. Overall, this suggests that the main gains come from the pairwise reward with Huber shaping, while explanation-based regularization provides at most a small additional benefit under our current design.

\subsection{WMT23 Metric Benchmark}

\noindent We report results on the WMT23 MQM metric shared task in Table~\ref{tab:wmt23}.
We compare Remedy-R against established reference-based metrics (e.g., XCOMET-XXL, MetricX-23), QE variants, and recent LLM-based judges.
For Remedy-R, we additionally evaluate test-time scaling (TTS) by sampling multiple reasoning outputs and aggregating their predicted scores (TTS=$k$).
We report both system-level accuracy (\textbf{Acc}) and segment-level \(\boldsymbol{\mathit{acc^*_{eq}}}\), together with the average of the two.\\

\noindent As shown in Table~\ref{tab:wmt23}, increasing TTS generally improves \(\boldsymbol{\mathit{acc^*_{eq}}}\) more consistently than \textbf{Acc}, leading to steady gains in the overall average.
This trend is most pronounced for smaller models (7B), where TTS closes a substantial portion of the gap to stronger baselines, and remains beneficial for 14B and 32B as well.
These results support our claim that sampling-based aggregation helps mitigate discrete score behaviors and yields more stable segment-level judgments.\\

\begin{table*}[h!]
\centering
\def\arraystretch{1.1}%
\resizebox{0.65\linewidth}{!}{%
\begin{tabular}{lcccc||c}
\toprule
\textbf{Method}& 
$\boldsymbol{\theta}$& 
\textbf{TTS/Turns} & 
\textbf{Acc} & 
\textbf{$\boldsymbol{\mathit{acc^*_{eq}}}$}
&
\textbf{Avg}
\\
 \midrule
KIWI-XXL & ensemble & 1 & 91.1 & 54.6 & 72.9 \\
MetricX-23 & 13B & 1 & 90.7 & 56.9 & 73.8 \\
MetricX-23-QE & 13B & 1 & 89.0 & 56.1 & 72.6 \\
XCOMET-XXL & ensemble & 1 & 92.8 & 57.7 & 75.3 \\
XCOMET-XXL-QE & ensemble & 1 & 91.6 & 55.8 & 73.7 \\
ReMedy$_\textnormal{9B-23}$ & 9B & 1 & 94.1 & 58.2 & 76.2 \\

\midrule
EAPrompt (GPT4o-mini) & - &  1 turn        & 90.3 & 45.9 & 68.1 \\
M-MAD (GPT4o-mini)    & - & 3 turns & 94.5 & 53.7 & 74.1 \\
GEMBA-MQM (GPT4) & -      & 1 turn   & 94.5 & 55.2 & 74.9 \\
\midrule
Remedy-R & 7B & 1 turn & 93.7  & 49.9 & 71.8 \\
Remedy-R & 7B & TTS=2 & 94.1  & 51.5 & 72.8 \\
Remedy-R & 7B & TTS=3 & 93.7  & 52.6 & 73.1 \\
Remedy-R & 7B & TTS=4 & 93.7  & 53.1 & 73.4 \\
Remedy-R & 7B & TTS=5 & 93.7  & 53.4 & 73.5 \\
Remedy-R & 7B & TTS=6 & 93.7  & 53.6 & 73.7 \\
\midrule

Remedy-R & 14B & 1 turn &  92.4 & 53.6 & 73.0 \\
Remedy-R & 14B & TTS=2 &  93.3 & 55.0 & 74.1 \\
Remedy-R & 14B & TTS=3 &  93.7 & 55.1 & 74.4 \\
Remedy-R & 14B & TTS=4 &  93.7 & 55.4 & 74.5 \\
Remedy-R & 14B & TTS=5 &  94.1 & 55.4 & 74.8 \\
Remedy-R & 14B & TTS=6 &  94.1 & 55.7 & 74.9 \\

\midrule
Remedy-R & 32B & 1 turn & 94.5 & 51.5 & 73.0 \\
Remedy-R & 32B & TTS=2 & 94.1 & 52.6 & 73.3 \\
Remedy-R & 32B & TTS=3 & 94.5 & 53.9 & 74.2 \\
Remedy-R & 32B & TTS=4 & 94.9 & 54.1 & 74.5 \\
Remedy-R & 32B & TTS=5 & 94.5 & 54.3 & 74.4 \\
Remedy-R & 32B & TTS=5 & 95.0 & 54.3 & 74.6 \\

\bottomrule 
\end{tabular}%
}
\caption{Evaluation on WMT MQM23 Metric Shared task. Both KIWI-XXL and XCOMET-XXL are identical ensembles of 2 $\times$ 10.7B and 1 $\times$ 3.5B models.}
\label{tab:wmt23}
\end{table*}

\subsection{WMT24 Metric Benchmark}

\noindent Table~\ref{tab:wmt24} reports results on the WMT24 MQM metric benchmark.
Following the official shared task protocol, we report system-level accuracy (\textbf{SPA}) and segment-level \(\boldsymbol{\mathit{acc^*_{eq}}}\), as well as their average correlation score (\textbf{Avg corr}) and the corresponding rank.\\

\noindent Remedy-R achieves competitive performance on WMT24 across model scales.
In particular, Remedy-R-14B ranks second overall and attains the best system-level accuracy (\textbf{SPA}).
While Remedy-R-32B and Remedy-R-7B show slightly lower segment-level \(\boldsymbol{\mathit{acc^*_{eq}}}\) than top baselines, their overall average remains comparable to strong reference-based metrics such as MetricX-24-Hybrid and XCOMET-XXL.\\

\begin{table}[h!]
\centering
\def\arraystretch{1.0}%
\setlength{\tabcolsep}{4pt}
\resizebox{0.5\linewidth}{!}{%
\begin{tabular}{lcccc}
\toprule
\multirow{2}{*}{\textbf{Methods}} & 
\multirow{2}{*}{\textbf{Rank}} & \multicolumn{1}{c}{\textbf{Avg}} & \multicolumn{1}{c}{\textbf{Sys}} & \multicolumn{1}{c}{\textbf{Seg}} \\
& & \multicolumn{1}{c}{\textbf{corr}} & \multicolumn{1}{c}{\textbf{SPA}} & \multicolumn{1}{c}{$\boldsymbol{\mathit{acc^*_{eq}}}$} \\
\midrule
ReMedy$_\textnormal{9B-24}$ & 1 & \textbf{72.9} & 85.9 & \textbf{60.0} \\
Remedy-R-14B & 2 & 72.6 & \textbf{87.9} & 57.2 \\


\midrule
MetricX-24-Hybrid & 3 & 72.1 & 85.6 & 58.5 \\
XCOMET-XXL & 4 & 71.9 & 86.1 & 57.6 \\

MetricX-24-Hybrid-QE & 5 & 71.4 & 84.9 & 58.0 \\
GEMBA-ESA (GPT4) & 6 & 71.1 & 84.6 & 57.6 \\
Remedy-R-32B & 7 & 70.6 & 85.3 & 56.2 \\
Remedy-R-7B & 8 & 70.6 & 85.8 & 55.4 \\
XCOMET-XXL-QE & 9 & 69.5 & 83.3 & 55.7 \\
Skywork-RM & 10 & 69.0 & 83.2 & 54.7 \\
\bottomrule 
\end{tabular}%
}
\caption{Evaluation on WMT24 MQM set. We report the official accuracy percentage (SPA and $\mathit{acc^*_{eq}}$).}
\label{tab:wmt24}
\end{table}

\subsection{GPT-4o-mini Faithfulness Prompt}\label{app:gpt4-faithfulness-prompt}

\noindent We use the following prompt to assess the faithfulness of Remedy-R’s evaluation explanations.
GPT-4 is given the source sentence, the translation hypothesis, and the explanation only (no reference translation), and returns a JSON object containing a \texttt{faithfulness\_score} and a brief reason.\\

\begin{Verbatim}[
  fontsize=\footnotesize,
  breaklines=true,
  breakanywhere=true,
  breaksymbolleft={}
]
System:
You are a strict verifier. Your job is to score the FAITHFULNESS of an evaluation explanation. You must return ONLY a single valid JSON object and nothing else.

User:
You are given: 1) src_sent: the source sentence; 2) target_sent: the translation hypothesis; 3) explanation: an evaluation text that comments on the translation quality

Task: Provide a score (0-100) indicating how FAITHFUL the explanation is to src_sent and target_sent.

Definition of faithfulness:
- Every key claim in the explanation must be supported by what is actually present in src_sent and/or target_sent.
- If the explanation invents content, mentions errors that are not evidenced, misquotes words, or contradicts src/target, score lower.

CRITICAL:
- You are NOT evaluating translation quality.
- A translation can be very bad, but an explanation can still be highly faithful if it correctly describes that badness.

Return ONLY JSON with:
{"faithfulness_score": int,   // 0-100
  "brief_reason": string       // <= 40 words, cite the biggest supported or unsupported claim}

Input:
src_lang: <src_lang>; tgt_lang: <tgt_lang>; src_sent: <src>; target_sent: <tgt>; explanation: <explanation>
\end{Verbatim}

\subsection{Remedy-R Agent experiments}
\noindent Table~\ref{tab:alma-r-double} reports additional Remedy-R Agent results on WMT24 using strong open-source ALMA-R translators as \(M_{\textit{base}}\).
We evaluate two base settings: ALMA-R-7B with Remedy-R-7B, and ALMA-R-13B with Remedy-R-14B.
For each language direction, we report multiple automatic metrics, including SacreBLEU, XCOMET-XXL, MetricX-24-XXL, and Remedy-R’s own scores.
We include both the initial translations (row ``- / -'') and the refined outputs produced by the Remedy-R Agent (row ``Remedy-R / Remedy-R''), where Remedy-R generates the explanation and performs the refinement based on its own feedback.\\

\noindent Overall, Remedy-R Agent consistently improves translation quality across language pairs, with particularly large gains on en-zh.
These results complement the main agent experiments and further demonstrate that Remedy-R’s reasoning can be reused to drive refinement even when the initial translations come from a strong external MT system.\\

\begin{table*}[h!]
\centering
\def\arraystretch{1.0}%
\setlength{\tabcolsep}{2pt}
\resizebox{\textwidth}{!}{%
\begin{tabular}{lllllll|lllllll}
\toprule
$M_\textit{feedback}$ & $M_\textit{refinement}$ & \textbf{en-cs} & \textbf{en-de} & \textbf{en-ru} & \textbf{en-zh} & \textbf{Avg} &
$M_\textit{feedback}$ & $M_\textit{refinement}$ & \textbf{en-cs} & \textbf{en-de} & \textbf{en-ru} & \textbf{en-zh} & \textbf{Avg} \\
\toprule

\multicolumn{14}{c}{\textbf{\textit{Base = ALMA-R-7B | Remedy-R = 7B}}} \\

\rowcolor[gray]{0.9}
\multicolumn{7}{c}{\textbf{SacreBLEU}} & \multicolumn{7}{c}{\textbf{XCOMET-XXL}} \\
- & - & 16.6 & 22.2 & 14.1 & 23.9 & 19.2 & - & - & 71.9 & 89.5 & 77.3 & 75.7 & 78.6 \\
Remedy-R & Remedy-R & 17.5\chg{+0.9} & 22.2\chg{0} & 15.8\chg{+1.7} & 27.4\chg{+3.5} & 20.7\chg{+1.5} &
Remedy-R & Remedy-R & 71.4\chg{-0.5} & 90.4\chg{+0.9} & 78.5\chg{+1.3} & 79.9\chg{+4.2} & 80.0\chg{+1.4} \\


\cmidrule(lr){1-7}\cmidrule(lr){8-14}
\rowcolor[gray]{0.9}
\multicolumn{7}{c}{\textbf{MetricX-24-XXL}} & \multicolumn{7}{c}{\textbf{Remedy-R}} \\
- & - &  -6.2 & -2.5 & -4.5 & -3.5 & -4.2 & - & - & 92.3 & 93.4 & 92.5 & 86.5 & 91.2 \\
Remedy-R & Remedy-R & -6.3\chg{+0.1} & -2.4\chg{+0.2} & -4.2\chg{+0.3} & -3.0\chg{+0.5} & -4.0\chg{+0.2} &
Remedy-R & Remedy-R & 94.2\chg{+1.9} & 95.4\chg{+2.0} & 94.6\chg{+2.1} & 92.0\chg{+5.5} & 94.1\chg{+2.9} \\

\midrule

\multicolumn{14}{c}{\textbf{\textit{Base = ALMA-R-13B | Remedy-R = 14B}}} \\
\rowcolor[gray]{0.9}
\multicolumn{7}{c}{\textbf{SacreBLEU}} & \multicolumn{7}{c}{\textbf{XCOMET-XXL}} \\
- & - & 18.2 & 22.8 & 15.9 & 26.2 & 20.8 &
- & - & 76.5 & 91.0 & 80.4 & 79.6 & 81.9 \\
Remedy-R & Remedy-R & 20.1\chg{+1.9} & 23.2\chg{+0.4} & 17.3\chg{+1.4} & 34.8\chg{+8.6} & 23.8\chg{+3.1} & Remedy-R & Remedy-R & 77.3\chg{+0.8} & 91.4\chg{+0.3} & 80.7\chg{+0.2} & 84.0\chg{+4.5} & 83.3\chg{+1.4} \\

\cmidrule(lr){1-7}\cmidrule(lr){8-14}
\rowcolor[gray]{0.9}
\multicolumn{7}{c}{\textbf{MetricX-24-XXL}} & \multicolumn{7}{c}{\textbf{Remedy-R}} \\
- & - & -5.2 & -2.1 & -3.8 & -3.1 & -3.6 &
- & - & 90.5 & 91.4 & 90.5 & 88.7 & 90.3 \\
Remedy-R & Remedy-R & -5.3\chg{-0.1} & -2.0\chg{+0.1} & -3.8\chg{+0.1} & -2.5\chg{+0.5} & -3.4\chg{+0.2} & Remedy-R & Remedy-R & 92.1\chg{+1.6} & 93.1\chg{+1.7} & 91.9\chg{+1.4} & 93.2\chg{+4.5} & 92.6\chg{+2.3} \\
\bottomrule
\end{tabular}%
}
\caption{Agent MT experiments on WMT24 using ALMA-R models as \(M_{\textit{base}}\).
We report initial translations (``- / -'') and Remedy-R Agent refinements (``Remedy-R / Remedy-R'') under multiple metrics.
Left: SacreBLEU. Right: XCOMET-XXL. We additionally report MetricX-24-XXL and Remedy-R scores in the lower blocks.}
\label{tab:alma-r-double}
\end{table*}

\end{document}